\documentclass[journal]{IEEEtran}
\usepackage{amsmath,amsfonts}
\usepackage{algorithmic}
\usepackage{algorithm}
\usepackage{array}
\usepackage[caption=false,font=normalsize,labelfont=sf,textfont=sf]{subfig}
\usepackage{textcomp}
\usepackage{stfloats}
\usepackage{url}
\usepackage{verbatim}
\usepackage{graphicx}
\usepackage{cite}
\usepackage{hyperref}
\usepackage{makecell}
\usepackage{multirow}
\usepackage{diagbox}
\usepackage{booktabs}
\usepackage{tablefootnote}
\usepackage{threeparttable}
\usepackage{graphicx}
\usepackage{tikz}

\newcommand\submittedtext{%
  \footnotesize This work has been submitted to the IEEE for possible publication. Copyright may be transferred without notice, after which this version may no longer be accessible.}

\newcommand\submittednotice{%
\begin{tikzpicture}[remember picture,overlay]
\node[anchor=south,yshift=10pt] at (current page.south) {\fbox{\parbox{\dimexpr0.65\textwidth-\fboxsep-\fboxrule\relax}{\submittedtext}}};
\end{tikzpicture}%
}

\begin{document}

\title{Hugging Rain Man: A Novel Facial Action Units Dataset for Analyzing Atypical Facial Expressions in Children with Autism Spectrum Disorder}

\author{Yanfeng Ji, Shutong Wang, Ruyi Xu, Jingying Chen$^*$, Xinzhou Jiang, Zhengyu Deng, Yuxuan Quan, Junpeng Liu 
\thanks{$^*$Corresponding author.}
\thanks{Yanfeng Ji, Shutong Wang, Jingying Chen, Xinzhou Jiang, Zhengyu Deng, Yuxuan Quan and Junpeng Liu are with the Faculty of Artificial Intelligence in Education, Central China Normal University, Wuhan, China, 430079.}
\thanks{Ruyi Xu is with the Computer Science and Artificial Intelligence School, Wuhan University of Technology, Wuhan, China, 430070.}
\thanks{E-Mail: \{jiyanfeng, wst\_2001, xinzhoun, zhengyudeng, yuxuanquan, ekails\}@mails.ccnu.edu.cn,  chenjy@mail.ccnu.edu.cn, ruyi.xu@whut.edu.cn}}

\markboth{Journal of \LaTeX\ Class Files,~Vol.~14, No.~8, August~2021}%
{Shell \MakeLowercase{\textit{et al.}}: A Sample Article Using IEEEtran.cls for IEEE Journals}


\maketitle
\submittednotice
\begin{abstract}
Children with Autism Spectrum Disorder (ASD) often exhibit atypical facial expressions. However, the specific objective facial features that underlie this subjective perception remain unclear. In this paper, we introduce a novel dataset, Hugging Rain Man (HRM), which includes facial action units (AUs) manually annotated by FACS experts for both children with ASD and typical development  (TD). The dataset comprises a rich collection of posed and spontaneous facial expressions, totaling approximately 130,000 frames, along with 22 AUs, 10 Action Descriptors (ADs), and atypicality ratings. A statistical analysis of static images from the HRM reveals significant differences between the ASD and TD groups across multiple AUs and ADs when displaying the same emotional expressions, confirming that participants with ASD tend to demonstrate more irregular and diverse expression patterns. Subsequently, a temporal regression method was presented to analyze atypicality of dynamic sequences, thereby bridging the gap between subjective perception and objective facial characteristics. Furthermore, baseline results for AU detection are provided for future research reference. This work not only contributes to our understanding of the unique facial expression characteristics associated with ASD but also provides potential tools for ASD early screening. Portions of the dataset, features, and pretrained models are accessible at: \url{https://github.com/Jonas-DL/Hugging-Rain-Man}.
\end{abstract}


\section{Introduction}
\IEEEPARstart{A}{utism} Spectrum Disorder (ASD) is a neurodevelopmental condition characterized by impairments in social interaction and the repetitive behaviors \cite{Lord2020AutismSD}. Facial expressions, serving as vital channels for conveying emotions and social intents, play a crucial role in social interactions \cite{Barrett2019EmotionalER}. Children with ASD often exhibit atypical, abnormal, or even discomfort-inducing facial expressions, which impede their social interactions and exacerbate their experiences of social isolation in daily encounters \cite{Briot2021NewTA,Grossard2020ChildrenWA}. These atypical facial expressions have also been extensively discussed as potential biomarkers for early screening of ASD \cite{Carpenter2020DigitalBP,Guha2015OnQF}. Therefore, investigating facial expressions in children with ASD holds paramount importance for enhancing our understanding of the social interaction impairments, as well as for developing screening methods for ASD \cite{Chen2021TowardCE,Perochon2023EarlyDO}.

Research into facial expressions in ASD predominantly relies on two annotation methodologies: basic facial expression labeling \cite{ekman1970universal} and the Facial Action Coding System (FACS) \cite{ekman1978facial}. Basic facial expressions are defined as universally recognized facial movements that convey distinct emotional states, such as happiness, sadness, anger, fear, disgust and surprise. By utilizing basic facial expression labeling, researchers can efficiently assess the ability of individuals with ASD to produce facial expressions that are congruent in response to their immediate emotional contexts. For example, Chen et al. found that \cite{Chen2021TowardCE}, compared to their TD peers, children with ASD exhibited predominant facial expressions that were less congruent with the emotional stimuli. However, when analyzing atypical facial expressions in individuals with ASD, basic facial expression labeling reveals certain limitations. Specifically, it lacks the necessary granularity to comprehensively characterize abnormal facial muscle movements and encounters difficulties in depicting ambiguous facial expressions \cite{Yirmiya1989FacialEO} and inferring their intended meanings. 

The Facial Action Coding System (FACS) \cite{Ekman2005WhatTF}, developed by psychologist Paul Ekman, decomposes facial movements elicited by expressions into Action Units (AUs) and Action Descriptors (ADs). 
The FACS offers a more granular, anatomically-based framework for analyzing atypical facial expressions. 
By examining combinations of AUs/ADs, researchers can more effectively capture the subtleties of facial expressions, as specific AUs may interact to convey complex emotions, which was emphasized by Du et al. study \cite{Du2014CompoundFE} on compound facial expressions. For instance, a genuine “Duchenne smile” involves both AU6 (cheek raise) and AU12 (lip corner pull), distinguishing it from an expression with AU12 alone. Furthermore, the temporal evolution of AUs/ADs is equally important; temporal analysis enables the capture of capture fleeting facial reactions, providing more nuanced emotional cues within a sequential context, as demonstrated in the works of Yang \cite{Yang2023TowardRF} and Tong \cite{Tong2007FacialAU} on robust AU detection integrating temporal information.

However, both annotating the basic categories of facial expressions and the composition of AUs are highly specialized, labor-intensive tasks. Leveraging advancements in computer vision technology, the development of facial expression recognition models and AU detection models offers promising avenues to enhance the efficiency of analyzing atypical facial expressions associated with ASD \cite{koehler2024machine,Drimalla2021ImitationAR,Zampella2020ComputerVA}. The cornerstone of training these models is the availability of large-scale, publicly accessible datasets. Unfortunately, the majority of automatic facial expression analysis tools \cite{Baltruaitis2018OpenFace2F,cheong2023py,bishay2023affdex}, which have primarily been trained on adult datasets \cite{Zhang2014BP4DSpontaneousAH,Mavadati2013DISFAAS,Lucey2010TheEC}, exhibit limitations in their generalization capabilities when tasked with analyzing children's facial expressions \cite{Witherow2024PilotST}. This phenomenon can be partially attributed to the significant physiological differences between adults and children, such as variations in facial structure, subcutaneous fat distribution, and wrinkle patterns \cite{ekman1978facial}. The most direct and effective way to address this issue is to train or fine-tune the model on datasets specifically curated for children.

The existing datasets of children's facial expressions present a multitude of challenges, such as data scarcity, restricted sample diversity, unprofessional annotations, and the absence of crucial labeling. As shown in Table \ref{table1}, several datasets that concentrate on the facial expressions of children with TD, notably CAFE \cite{Lobue2014TheCA}, ChildEFES \cite{Negro2021TheCE}, and LIRIS-CSE \cite{Khan2018AND}, merely offer basic expression category labels and fail to provide AU labels. When it comes to datasets focusing on ASD, the scarcity issue becomes even more acute \cite{Pandya2023ACA,deBelen2020ComputerVI}; publicly available datasets are limited to Kaggle-Autism\footnote{\href{https://www.kaggle.com/discussions/general/123978}{Kaggle-Autism}} and DASD. Both of these datasets fall short of the essential features  required for in-depth expression analysis: Kaggle-Autism provides only static facial images along with non-expert diagnostic labels, whereas DASD lacks raw video data and dependable manual annotations. Consequently, there is an urgent need to develop a new, large-scale dataset of facial expressions in individuals with ASD to facilitate the advancement of automatic facial expression annotation tools.

\begin{table*}[htbp]
\caption{Facial or Facial Expression Datasets of Children with TD and Children with ASD.\label{table1}}
\centering
\resizebox{\linewidth}{!}{

\begin{tabular}{cccccccc}
\toprule
Database & Participants & Ages & Sex(M:F) & Frames & Facial Expression Type& Environment & Annotation\\
\midrule
CAFE\cite{Lobue2014TheCA} & 154 & 2-8 &64:90 & 1,192&Posed &Lab&Natural + 6 basic expressions\\
NIMH-ChEF\cite{Egger2011TheNC} &482 & 10-17 &193:341 &482 & Posed & Lab&Natural + 4 basic expressions\\
LIRIS-CSE\cite{Khan2018AND} &12 & 6-12 &5:7 &208 segments & Spontaneous & Classroom/Lab/Home&6 basic expressions\\
Dartmouth\cite{Dalrymple2013TheDD} &80 & 6-16 &40:40 &- & Posed & Lab&6 basic expressions\\
ChildFES\cite{Negro2021TheCE} &132 & 4-6 &- &1,014 photos \& 971 videos & Posed  & Lab&Natural + 7 basic expressions\\
Kaggle-Autism &- & - &- &2,926 & - & -&Binary label\\
DASD\cite{Cai2022AnAD} &82 & - &ASD(57) TD(25) &41,002 & - & -&Openface features\\
HRM &98 & 2-12 &ASD(53:13) TD(18:14) &131,758|1,535 segments & Posed \& Spontaneous & Classroom&22 AUs + 10 ADs, Atypicality ratings\\
\bottomrule
\end{tabular}
}
\end{table*}

Beyond labeling facial expression categories or AUs/ADs, an additional critical challenge is the recognition of atypical facial expressions in individuals with ASD. Existing methods primarily focus on exploring suitable visual features to quantify the differences in facial expressions between individuals with ASD and their TD peers\cite{Guha2018ACS,Krishnappababu2021ExploringCO,Witherow2024PilotST}. However, such efforts frequently overlook the assessment of the severity of atypical facial expressions in ASD. In certain empirical studies, individuals with typical development are recruited to perceive and evaluate the facial expressions of those with ASD \cite{Brewer2015CanNI,faso2015evaluating,trevisan2018facial}. This approach facilitates the recognition of varying degrees of atypicality across different facial expressions within the ASD cohort. Consequently, a pivotal question arises for computer vision tasks: how can computational systems correlate human subjective perceptions of atypicality with specific facial action features? Addressing this question holds the potential not only to facilitate the deconstruction of facial expression patterns deemed 'atypical,' elucidating their structural characteristics and dynamic patterns, but also to establish a scientific foundation for an objective evaluation framework for facial expression atypicality. In turn, this could enhance our understanding of the social interaction challenges encountered by individuals with ASD.

To address these challenges, we construct the \textbf{H}ugging \textbf{R}ain \textbf{M}an (HRM) dataset, a novel facial action unit dataset. Named in homage to both the machine learning community Hugging Face and the Oscar-winning film \textit{Rain Man}, HRM symbolizes our commitment to using AI to decode emotional expression patterns of children with ASD and to fostering greater understanding and awareness of the ASD community. The HRM dataset includes facial expression data from from 66 children with ASD and 32 children with TD, totaling nearly 130,000 FACS-expert-annotated images with 22 AUs and 10 ADs. We evaluate multiple baseline AU detection models on the HRM dataset and reported their performance. Additionally, we introduce atypicality rating annotations and explore the feasibility of using AU temporal features to predict perceived atypicality, thereby providing a novel research avenue for quantifying and understanding atypical expressions in children with ASD.

The key contributions of this paper are as follows:

\begin{enumerate}
\item{To the best of our knowledge, this work presents the first dedicated dataset of facial action units for children with ASD, called the Hugging Rain Man (HRM) dataset. The dataset includes a substantial amount of facial expression data from children with ASD alongside control data from  children with TD. We constructed the HRM dataset using a multi-dimensional annotation framework that encompasses 22 AUs, 10 ADs, and atypicality ratings for expression segments. The structured collection and annotation of this dataset lay a foundation for in-depth comparative analysis of facial expression features specific to children with ASD.}
\item{We conducted an in-depth examination of the static facial AU differences between children with ASD and children with TD under identical emotional conditions. The results indicate that, compared to their TD peers, children with ASD exhibit significant differences in multiple AUs/ADs across three basic facial expressions: happiness, surprise, and sadness. Additionally, the ASD group shows higher AU combination complexity and greater diversity in AU combination types. This fine-grained AU-based analytical method not only provides a new perspective for understanding emotional expressions in children with ASD but also offers potential biomarkers for early screening of ASD.}
\item{We employed a temporal regression model to explore the relationship between objective FACS metrics and the perceived atypicality of facial expressions, to quantitatively analyze dynamic facial expressions in children with ASD. By conducting a fine-grained analysis of AU sequences, we explored how temporal patterns in expression generation contribute to perceived atypicality, creating a bridge between subjective perception and objective facial characteristics at the "perception-mechanism" level. This temporal feature-based quantitative analysis not only helps deconstruct and understand the objective features leading to perceived atypicality in expressions but also introduces a novel quantitative approach for researching facial expressions in children with ASD.}
\item{We trained and evaluated multiple mainstream AU detection algorithms on the HRM dataset and made these optimized pre-trained models available to the research community. This provides other researchers with more reliable tools, which are expected to accelerate the development and application of facial expression analysis technology for children, particularly those with ASD.}
\end{enumerate}

The rest of this paper is organized as follows. Section \ref{sec2} provides a summary of existing face datasets related to children, discussing prior research that utilized AU analysis to study the facial expressions of  children with ASD. Section \ref{sec3} details the construction of the dataset, including participant selection, experimental design, data annotation, and consistency checks. Section \ref{sec4} analyze the characteristics of the dataset and examine the differences in AUs and ADs for three basic facial expressions between the two groups. Section \ref{sec5} establishes baseline models for the atypicality ratings regression and AU detection. Section \ref{sec_dicus} discusses issues related to experimental results and experimental paradigms. Finally, Section \ref{sec_conclu} concludes the paper and outlines future directions for research.

\section{Related Work\label{sec2}}
In this section, we first introduce face datasets related to children, including both children with TD and children with ASD, and then investigate the differences in AUs between individuals with ASD and their TD peers under the same emotional state.
\subsection{Children's Face Dataset}
\subsubsection{\textbf{TD Dataset}}
The CAFE dataset \cite{Lobue2014TheCA} employs an expression imitation paradigm, collecting images of six basic emotions (happiness, anger, fear, sadness, disgust, and surprise) along with neutral faces from 100 children aged 2 to 8 years in a laboratory setting, resulting in a total of 1,192 frames. Annotation was performed by a FACS-trained expert in collaboration with 100 untrained participants. Although the sample size is relatively small, the dataset boasts considerable ethnic diversity (17\% African American, 27\% Asian, 30\% White, 17\% Latino, and 9\% “Other”). Access to this dataset requires official authorization.

The NIMH-ChEF dataset \cite{Egger2011TheNC} consists of 482 images labeled with four basic emotions (happiness, fear, anger, and sadness) as well as neutral faces. Its free release provides high-quality image material for analyzing children's responses to peer facial expression stimuli.

The Dartmouth dataset \cite{Dalrymple2013TheDD} induces children to imagine scenarios that evoke specific emotions through verbal prompts, capturing facial expressions from five different angles. The study recruited 80 individuals aged 6 to 16 and labeled six basic emotions along with their respective intensity levels (ranging from 1 to 5), adding an additional label “None” for ambiguous expressions. Image acquisition requires an application submission.

The ChildFES dataset \cite{Negro2021TheCE} includes facial expression videos of 132 children aged 4 to 6, featuring seven basic emotions (adding “contempt” to the six basic emotions) and neutral faces. The data was collected using both imitation and guided imagination techniques, and access requires application.

Most of the aforementioned datasets primarily focus on “posed” expressions, which may obscure children's genuine emotional expressions. In contrast, the LIRIS-CSE dataset \cite{Khan2018AND} collects spontaneous facial expression videos from 12 children aged 6 to 12 from diverse ethnic backgrounds, encompassing a total of 208 clips captured in natural settings, including classrooms, labs, and home environments, offering a more authentic representation of emotional expression.

\subsubsection{\textbf{ASD Dataset}}
Currently, there are only two open-source face or facial expression datasets related to ASD: Kaggle-Autism and DASD \cite{Cai2022AnAD}. The scarcity of such datasets is primarily due to significant privacy issues and ethical considerations associated with facial data, making the collection and sharing of these datasets challenging.

The Kaggle-Autism dataset consists of 2,926 facial images of individuals with ASD and children with TD sourced from the internet. It is important to note that this dataset lacks expert diagnostic records and relies solely on keyword searches for image retrieval.

The DASD dataset includes data from 57 individuals with ASD and 25 individuals with TD, where parents captured facial data by making simple changes to the angle of their mobile phone cameras. OpenFace 2.0 \cite{Baltruaitis2018OpenFace2F} was utilized to extract various anonymized features related to head pose, facial landmarks, and AUs. While the DASD dataset protects participants' privacy through anonymized features, it suffers from the loss of original image information and lacks precise AU annotations.

Overall, existing datasets predominantly involve laboratory settings, lacking genuine emotional expressions in natural contexts. Emotion elicitation methods mainly rely on imitation and verbal prompts, resulting in a scarcity of spontaneous emotional displays. In terms of labeling, basic facial expressions are the standard in most studies, and apart from DASD, there is a lack of detailed AU annotations, with DASD relying on machine-generated pseudo-labels. Additionally, many datasets have limited sample sizes, hindering effective training of deep learning models. Finally, only DASD includes both participants with ASD and those with TD. Consequently, the existing datasets do not comprehensively reflect the facial expression characteristics and dynamics of children, thereby restricting the depth and breadth of related research.

\subsection{Analysis of Facial Action Units in Children with ASD}
Happiness is the preferred emotion for analyzing facial expressions in individuals with ASD. Associated AUs typically include AU6 and AU12. This emotion is easily elicited and aligns with ethical considerations. Abigail and colleagues utilized humorous videos as stimuli in their research \cite{Bangerter2020AutomatedRO}, analyzing the differences in AU features related to happiness between ASD and TD groups using the FACET tool. Research findings indicate that, on average, the ASD group exhibits relatively less activity in AU6 and AU12 during happy expressions compared to the TD group. Similar observations have been reported in studies \cite{Manfredonia2018AutomaticRO,Samad2019APS,Zampella2020ComputerVA,Alvari2021IsST}. Witherow et al. also found that the ASD group produced significantly fewer instances of AU6 when imitating happy expressions compared to the control group \cite{Witherow2024PilotST}. Additionally, temporal analysis of AU activation revealed a low correlation between the combinations of AU6+AU12 and AU10+AU12 in the ASD group \cite{Samad2019APS}. Other studies have noted that individuals with ASD may exhibit uncontrolled occurrences of AU12, which do not align with their visual engagement (eye movements) while viewing 3D facial expressions \cite{Samad2018AFS}. This spontaneous smiling, lacking appropriate visual engagement, suggests potential barriers for individuals with ASD in achieving reciprocal communication.

Surprise is often regarded as the purest emotion, neither sad nor happy, and it tends to be fleeting, quickly transitioning to another emotion. Consequently, what we often observe is a blend of surprise with other emotions. The action units that typically constitute a surprised expression include AU1, AU2, AU5, AU25, AU26, and AU27. Research indicates significant differences between the ASD and TD groups regarding surprise (AU5) \cite{Manfredonia2018AutomaticRO}, with the former exhibiting higher activation intensity in AU2 when imitating surprised expressions \cite{Witherow2024PilotST}.

Negative emotions generally encompass anger, disgust, sadness, and fear. significant differences were reported between the ASD and TD groups in expressions of fear (AU5) and disgust (AU9) \cite{Manfredonia2018AutomaticRO,Witherow2024PilotST}, while differences in expressions of sadness or anger were not as pronounced. However, in another study \cite{Samad2019APS}, it was indicated that individuals with ASD displayed atypical activation of AU15 (lip corner depressor), which is generally associated with sadness, in response to excited, disgusted, and tense expressions presented by a 3D avatar. This research employed commercial facial capture software, Faceshift\footnote{\url{https://www.venturelab.swiss/faceshift}}, to capture detailed facial expressions, including 10 AUs. The findings also highlighted significant differences in AU20, which typically appears in fearful expressions, between the ASD and TD groups.

Individuals with ASD typically exhibit poorer facial symmetry \cite{Guha2015OnQF,Grossard2020ChildrenWA}. Research \cite{Witherow2024PilotST} has revealed that the ASD group displays greater asymmetry in AU5 and AU25. Samad et al. found significant differences in muscle responses responsible for AU13 between the left and right sides within the ASD group \cite{Samad2015AnalysisOF}. This suggests that facial muscle asymmetry may contribute to the more peculiar expressions observed in individuals with ASD.

Manual AU annotation for participants with ASD is not common. The work by Weiss et al. \cite{Weiss2019LessDF} is currently the only known work that provides comprehensive AU annotations for emotional stimuli and peak frames of recorded videos. Analyzing more complex facial muscle response patterns revealed differences between individuals with high-functioning ASD (HF-ASD) and individuals with TD. Compared to individuals with TD, individuals with HF-ASD  more frequently activated muscle movements that did not align with the AUs depicted in the stimuli.

Most of the aforementioned studies have utilized commercial or open-source AU detection/intensity estimation tools, such as OpenFace, FACET, Faceshift, and Imotions AffDex. These tools significantly reduce the time cost of AU annotation due to their ease of deployment, support for real-time processing, and automation capabilities. However, it's crucial to consider that the algorithms of these tools are primarily trained on adult facial datasets (e.g., BP4D \cite{Zhang2014BP4DSpontaneousAH}, DISFA \cite{Mavadati2013DISFAAS}, CK+ \cite{Lucey2010TheEC}). Given the uniqueness and heterogeneity of ASD facial expressions \cite{Trevisan2018FacialEP}, we are concerned about their accuracy and reliability when applied to children and adolescents with ASD.

\section{Description of the HRM Dataset\label{sec3}}
This section describes the participants involved in the creation of the HRM dataset, the experimental tasks, the data annotation process, and the analysis of annotation reliability.

\subsection{Participants}
This study was conducted following approval from the Ethics Committee of Central China Normal University (Approval No. CCNU-IRB-202312043b). Two capability assessment experiments were rigorously implemented at a special education school and a regular kindergarten. All caregivers of the participating children will be provided with informed consent forms after being informed about the experimental tasks.

In the informed consent form, we outlined the purpose and methods of the research, informing participants of their right to withdraw from the study at any time. The experiments were carried out in the special education school and the regular kindergarten, with each session lasting approximately 10 to 30 minutes. During the experiments, children engaged in a series of games and simple interactive tasks. We utilized various devices to collect multimodal data generated by the children during the activities. To protect the rights of participants, all collected data will be kept strictly confidential, and personal information will not be disclosed. The data will be used solely for research purposes and stored on encrypted electronic devices. Additionally, each child participating in the experiments will receive a detailed assessment report, which will include their performance during the experiment and relevant recommendations. All relevant entries will be explained in detail to the caregivers, who will sign the informed consent form after fully understanding and agreeing to the terms.

The children with ASD recruited from the special education school all met the DSM-5 criteria \cite{regier2013dsm}. A total of 140 children aged 2-12 years were recruited (ASD = 78, TD = 62). All  children with ASD participating in the two ability assessment experiments also took the Peabody Picture Vocabulary Test \cite{dumont2008peabody,Krasileva2017PeabodyPV}. We selected 66 children with ASD (53 male, 13 female) and 32  children with TD (18 male, 14 female) from the total participants for data annotation, detailed information is provided in Table \ref{table2}.

\begin{table}[!t]
\caption{Sample Demographics\label{table2}}
\centering
\begin{tabular}{ccc}
\toprule
  & ASD (N=66) & TD (N=32)\\
\midrule
Age & 5.29$\pm$2.30& 4.37$\pm$1.62\\
Male & 53 (80\%) & 18 (56\%)\\
Female & 13 (20\%) & 14 (44\%) \\
Verbal IQ & 55.18$\pm$28.42 & n/a\\ 
\bottomrule
\end{tabular}
\end{table}

\subsection{Experimental Settings}
The children's facial videos used in this dataset are partially derived from a previously published study by our research group \cite{Chen2020AnIM,Chen2021TowardCE}, with the remaining videos collected during two ability assessment experiments conducted at special education schools and kindergartens from December 2023. Studies \cite{Chen2020AnIM,Chen2021TowardCE} aimed to elicit spontaneous facial expressions from participants by presenting visual stimuli that contained both social and non-social elements. The ability assessment experiments consisted of two main tasks: the first focused on facial expression recognition and imitation, designed to evaluate participants' emotional perception and ability to replicate expressions, as shown in Fig.1a-1b. The second task, a graphical analogy reasoning test, aimed to assess participants' cognitive abilities in abstract thinking, problem-solving, and logical reasoning, as illustrated in Fig.1c-1d. Below, we provide a brief introduction to the experimental setup and the design of the ability assessment tasks.

\begin{figure}[!t]
\centering
\includegraphics[width=0.45\textwidth]{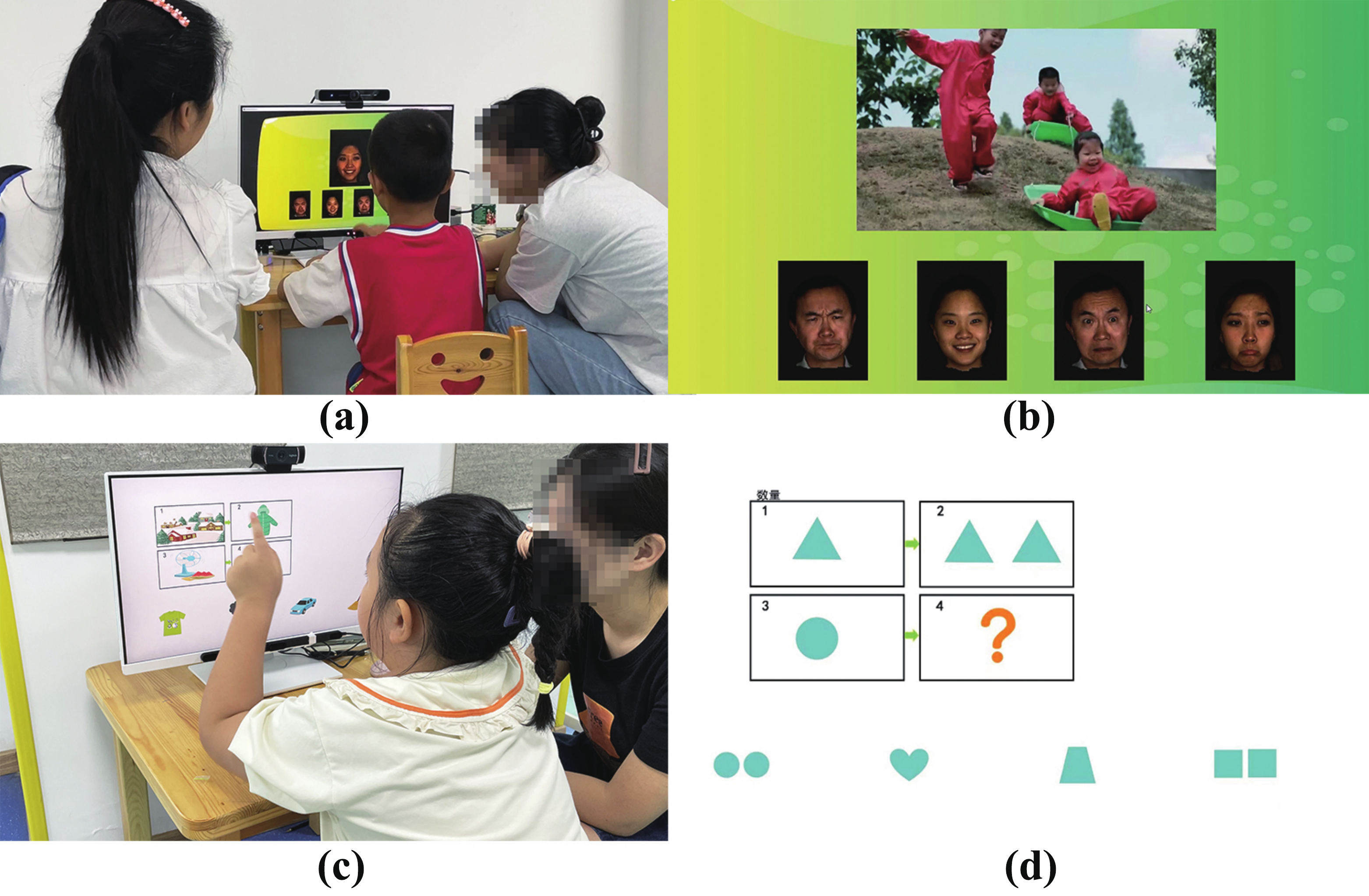}

\caption{Facial expression recognition, imitation and induction, as well as figural analogy reasoning experiments. (a) and (b) refer to static and dynamic facial expression recognition, imitation and induction games, respectively. (c) and (d) Logical reasoning games.}
\label{fig_game}
\end{figure}

\textbf{Equipment:} The experiment was conducted using a Philips Curio 24-inch all-in-one computer with a screen resolution of 1920$\times$1080. The eye-tracking data were collected using a Tobii Eye Tracker 5, and skin conductance was recorded with an Empatica E4 wristband Rev. 2. The video recordings were captured using a Hikvision 4K camera and a Logitech 1080P camera.

\textbf{Experimental Tasks:}
\begin{itemize}
\item{\textbf{Facial Expression Recognition, Imitation and Induction Task:} This task consists of two games: static and dynamic facial expression game. In the static expression game, children are shown various facial expression images selected from the BU-4DFE dataset \cite{Yin2008AH3}. They are asked to identify the emotion represented by the image and select the corresponding expression from the options displayed at the bottom of the screen. After each selection, the program automatically records the score, and the experimenter asks the child: "Can you try to imitate this expression?" In the dynamic expression game, children watch a video portraying a real-life scenario in which a person expresses a specific emotion. After viewing the video, the child is asked to select the correct emotion from four options. Once a choice is made, the experimenter asks: "When you feel happy (or the emotion shown in the scene), what kind of facial expression would you make?" In this task, children may produce posed and spontaneous facial expressions.}
\item{\textbf{Graphic Analogy Reasoning Test:} This task presents children with a series of analogy problems related to everyday life. The child must select the most logically fitting figure from four options based on the experimenter’s prompt. For example, as shown in Fig. 1d, the game presents a problem involving the transformation of triangle quantities, and the child is required to deduce the analogous change in circle quantities. If the child is unable to complete the task independently, the experimenter provides a hint, such as: "One triangle turns into two triangles, so what should one circle turn into?" In this task, children will produce various spontaneous facial expressions, such as happiness, confusion, etc.}
\item{All games were developed using Unity.}
\end{itemize}

\subsection{FACS Annotation}
\begin{figure*}[!t]
\centering
\includegraphics[width=0.95\textwidth]{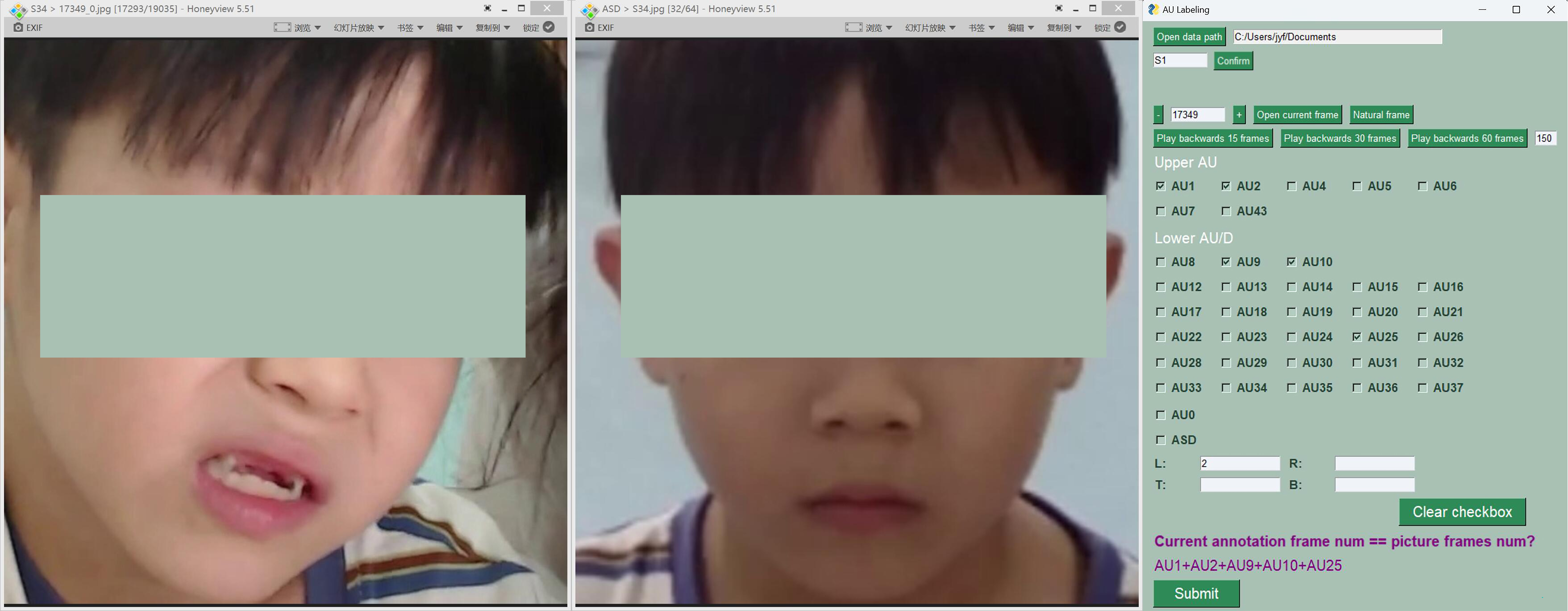}
\caption{AU/AD Annotation Tool. The left and middle positions show the children's expression frames and neutral frames, respectively, with the eye area covered in green-gray to ensure privacy. The far right shows the developed AU/AD annotation tool.}
\label{au_lableing}
\end{figure*}
Due to individual differences in cognitive abilities and IQ, the duration each child took to complete different experiments varied. Additionally, as participants came from multiple institutions and did not undergo identical experimental procedures, this dataset is organized by participant rather than by specific task paradigms, unlike the BP4D dataset \cite{Zhang2014BP4DSpontaneousAH}. Although some children were unable to complete the entire experimental process, their facial expressions were still retained in the dataset. Participants with less than one minute of recorded video were excluded from the dataset.

\textbf{Clip Selection:} Experts reviewed each participant’s video and selected segments with the richest facial expressions, particularly those that appeared atypical or awkward, for annotation. The selected clips typically began with a neutral expression and continued until the expression concluded. If a child left the frame midway or if their head movements exceeded the annotatable range, those segments were discarded.

\textbf{Face Detection:} To focus solely on the face rather than the entire frame, we used RetinaFace \cite{Deng2019RetinaFaceSD} for face localization. The detected faces were resized to a uniform 512$\times$512 resolution. Facial alignment was intentionally not performed to preserve the consistency of the facial expressions. This choice ensures that subtle nuances in expression across frames are maintained, especially since facial misalignment might obscure key expression-related features, such as the movement of smaller muscles. By keeping the raw orientation of the face, we can better analyze expressions that involve asymmetrical movements, which are particularly relevant for ASD studies.

\textbf{Annotation Tool and Rules:} We developed a simple frame-level AU annotation tool using the PySimpleGUI library\footnote{ \url{https://pypi.org/project/PySimpleGUI/}}, as shown in Fig.\ref{au_lableing}. Experts could select the participant to be annotated, and simultaneously use HoneyView to display both the frame to be annotated and the participant's neutral frame. When annotating a specific frame, experts could navigate through previous and subsequent frames using the left and right arrow keys. The tool also supports video playback from the current frame at a custom speed, allowing experts to closely observe facial muscle movements. Considering that existing research suggests children with ASD exhibit more asymmetrical facial expressions compared to  children with TD \cite{Guha2018ACS}, the tool has been enhanced with the ability to distinguish between left, right, up, and down, using identifiers 2, 3, 4, and 5, respectively.
It is important to note that, for simplicity, we did not explicitly distinguish between an active jaw drop (AU27) and a passive jaw drop caused by muscle relaxation (AU26). Instead, we used a unified label, AU2X, to describe the action of "jaw dropping," which encompasses both AU26 and AU27.

The entire dataset was fully annotated by the first expert, while the second expert randomly selected participants for annotation to perform inter-rater reliability analysis. To minimize the risk of missing any AU/AD annotations, all AUs/ADs were listed as checkboxes, and experts were required to review each one during the annotation process. Before submitting the final labels, experts were instructed to double-check both the checkboxes and the results, following the FACS manual’s guideline \cite{ekman1978facial}: “Never try to do the Omission Check in one viewing!” On average, approximately 1,500 frames were annotated for each  participant with ASD, while each  participant with TD had about 900 frames annotated.

\subsection{Annotation Reliability\label{FACS reliabi}}
Open-source and commercial tools that support AU detection or intensity estimation, such as OpenFace 2.0 \cite{Baltruaitis2018OpenFace2F}, Py-feat \cite{cheong2023py}, AFFDEX 2.0 \cite{bishay2023affdex}, AFAR \cite{Ertugrul2019CrossdomainAD}, and FaceReader\footnote{\url{https://www.noldus.com/facereader}}, have garnered significant attention from researchers. These tools are widely used in fields like psychology \cite{Monaro2021DetectingDT,Stckli2018FacialEA} and medicine \cite{Glen2023AIASSISTEDEA,Koehler2022MachineLC} due to their ease of deployment, real-time processing capabilities, and the ability to automate processes, which significantly reduces the time cost of AU annotation. Before considering manual AU annotation, we also turned our attention to these open-source tools, hoping to leverage them for analyzing facial AUs in children with ASD. However, it is important to exercise caution, as these tools are primarily trained on adult facial datasets (e.g., BP4D \cite{Zhang2014BP4DSpontaneousAH}, DISFA \cite{Mavadati2013DISFAAS}, and GFT \cite{Girard2017SayetteGF}), and known AU labeling biases \cite{Tan2024CausallyUB,Cui2020LabelEC} in the training data raise concerns about their accuracy and reliability when applied to the unique and sensitive context of children with autism. To address these concerns, we selected two open-source AU detection tools, OpenFace 2.0 and Py-feat, to assess their adaptability and robustness in the specific context of AU detection for children with autism.

We first randomly selected seven participants from the ASD group. Two  FACS-certified experts (one male and one female) used a custom annotation tool to identify the most expressive video segments and annotate 33 AUs/ADs, as shown in Fig. \ref{au_lableing}. Each participant had approximately 1,500 frames labeled, resulting in a total of around 10,000 annotated frames. When head movements in the video exceeded the visible range, making accurate AU annotation difficult, we carefully excluded these segments to ensure the precision and consistency of the annotations.

Next, the video data for each participant were processed using OpenFace and Py-feat. OpenFace utilized a dynamic AU model, while Py-feat employed an Support Vector Machine (SVM)-based model \cite{hearst1998support} for AU detection. We then selected the frames corresponding to the expert-labeled ones to calculate consistency between the expert annotations and the tool outputs. It's important to note that the two experts distinguished between different regions (left, right, upper, and lower) when annotating AUs, using codes 2, 3, 4, and 5, while 0 and 1 indicated non-activation and bilateral activation, respectively. Therefore, labels greater than 1 were converted to 1 for comparison.

\begin{table*}[!htbp]
\caption{Action Units Description, Distribution and Reliability Comparison\label{au_dist_reli}}
\centering
\resizebox{\linewidth}{!}{
\begin{threeparttable}
\begin{tabular}{ccccccccccc}
\toprule
AU/AD & Name & ICC(A-B) & ICC(A-O) &ICC(B-O)& ICC(A-P) & ICC(B-P)&F1(A-O) & F1(A-P) & AU OCC \\
\midrule
1 & Inner Brow Raiser & 0.680 & 0.270 & 0.209&0.324&0.286&0.416 & 0.449&17,839\\

2 & Outer Brow Raiser & 0.862 & 0.049 & 0.062 & 0.121 & 0.145&0.199 & 0.257&14,118\\

4 & Brow Lowerer & 0.875 & 0.201 & 0.209 & 0.261 & 0.262&0.414 & 0.422&13,988\\

5 &Upper Lid Raiser& 0.946 & 0.053 & 0.047 & 0.031 & 0.032&0.050 & 0.033&524\\

6&Cheek Raiser and Lid Compressor&0.894&0.470&0.392&0.457&0.381&0.511 & 0.486&12,545\\

7     & Lid Tightener          & 0.757 & 0.239  & 0.247  & 0.287  & 0.305  &0.326 & 0.399&  19,733\\
8     & Lips Toward Each Other  & 0.733 & n/a      & n/a      & n/a      & n/a      &n/a & n/a& 599 \\
9     & Nose Wrinkler          & 0.816 & 0.143  & 0.148  & 0.241  & 0.258  &0.168 & 0.279& 1,924 \\
10    & Upper Lip Raiser       & 0.714 & 0.087  & 0.076  & 0.089  & 0.102 & 0.105& 0.114 & 6,924 \\
12    & Lip Corner Puller      & 0.746 & 0.485  & 0.416  & 0.450  & 0.403  &0.582 & 0.526& 31,623 \\
13    & Sharp Lip Puller       & n/a   & n/a      & n/a      &n/a     & n/a    &n/a & n/a& 184\\
14    & Dimpler                & 0.794 & 0.254  & 0.209  & 0.211  & 0.167  &0.437 & 0.406& 18,937 \\
15    & Lip Corner Depressor   & 0.758 & -0.003 & -0.005 & -0.002 & 0.002  &0 & 0&684  \\
16    & Lower Lip Depressor    & 0.762 & n/a      & n/a      & n/a      & n/a      &n/a & n/a& 7,586 \\
17    & Chin Raiser            & 0.757 & 0.222  & 0.216  & 0.021  & -0.003 &0.288 & 0.119& 6,386 \\
18    & Lip Pucker             & 0.719 & n/a      & n/a      & n/a      & n/a      & n/a & n/a&3,440 \\
19    & Tongue Show            & 0.945 & n/a      & n/a      & n/a      & n/a      &  n/a & n/a&2,772\\
20    & Lip Stretcher          & 0.795 & 0.141  & 0.109  & 0.091  & 0.085  &0.280 & 0.198& 3,870 \\
22    & Lip Funneler           & 0.665 & n/a      & n/a      & n/a      & n/a      &n/a & n/a& 1,760 \\
23    & Lip Tightener          & 0.913 & 0.008  & 0.020  & 0.021  & 0.026  &0.084 & 0.163& 6,090 \\
24    & Lip Presser            & 0.886 & n/a      & n/a      & 0.094  & 0.100  &n/a & 0.130& 5,968\\
25    & Lips Part              & 0.932 & 0.417  & 0.405  & 0.364  & 0.368  &0.750 & 0.685& 80,683\\
26/27 & Jaw Drop/Mouth Stretch & 0.969 & 0.295  & 0.284  & 0.378  & 0.367  & 0.520 & 0.516& 51,182\\
28    & Lip Sucker             & 0.816 & 0.152  & -0.011 & 0.024  & 0.023  &0.162 & 0.035& 3,840 \\
29    & Jaw Thrust             & 0.937 & n/a      & n/a      & n/a      & n/a      &n/a & n/a& 816 \\
30    & Jaw Sideways           & 0.748 & n/a      & n/a      & n/a      & n/a      & n/a & n/a&1,796 \\
32    & Bite                   & 0.970 & n/a      & n/a      & n/a      & n/a      &n/a & n/a& 2,038\\
33    & Blow                   & n/a   & n/a      & n/a      & n/a      & n/a      &n/a & n/a& 31\\
34    & Puff                   & n/a   & n/a      & n/a      & n/a      & n/a      & n/a & n/a& 696\\
35    & Suck                   & n/a   & n/a      & n/a      & n/a      & n/a      & n/a & n/a&277 \\
36    & Bulge                  & n/a   & n/a      & n/a      & n/a      & n/a      &n/a & n/a& 63\\
37    & Lip Wipe               & 0.842 & n/a      & n/a      & n/a      & n/a      & n/a & n/a& 783\\
43E   & Eye Closure            & 0.962 & n/a      & n/a      & 0.068  & 0.061  &n/a & 0.073& 3,303\\
Avg   &                        & 0.859 & 0.205  & 0.178  & 0.185  & 0.177&0.311 & 0.278&  \\
\bottomrule
\end{tabular}
\begin{tablenotes}
        \footnotesize
        \item[1] The letters A and B in parentheses represent Expert A and Expert B, respectively. O stands for Openface, P stands for Pyfeat, and OCC stands for 'occurrence'.
      \end{tablenotes}
\end{threeparttable}
}
\end{table*}
Following the protocols of DISFA and RFAU \cite{Hu2020RFAUAD}, inter-rater reliability between the two experts was assessed using the two-way mixed-effects model \cite{Koo2016AGO}, ICC (3,1). An ICC value above 0.8 is considered high reliability, while values below 0.5 indicate low reliability. As shown in Table \ref{au_dist_reli}, the ICC values between the two experts ranged from 0.66 to 0.97. However, the two open-source tools achieved only moderate consistency (ICC around 0.4) for AU6, AU12, and AU25, with poor agreement on other AUs. This is strongly related to the imbalanced distribution of AU labels in their training datasets. For example, AU25 accounts for approximately 30\% of the samples in the DISFA dataset \cite{Ma2022FacialAU}, while AU6 and AU12 are much more prevalent in BP4D and BP4D+ datasets \cite{Zhang2016MultimodalSE}. Training on highly imbalanced datasets leads to model biases toward majority classes.

Finally, we used the annotations from Expert A as the ground truth and calculated the F1-score of the predictions generated by OpenFace and PyFeat. This metric is commonly used to evaluate model performance in AU detection tasks. The results revealed that the average F1-score for both open-source tools was only around 0.3, indicating that the current automated AU detection tools have limited effectiveness in analyzing facial expressions of children with ASD.

In summary, while existing open-source AU detection tools show potential and convenience, their feasibility and effectiveness in accurately detecting AUs in children with ASD face significant challenges. These challenges stem from the heavy reliance on adult data during tool development, the inherent biases in the training data, and the failure to account for the unique and heterogeneous facial expressions of children with ASD \cite{Carpenter2020DigitalBP}. Therefore, it is crucial and urgent to develop child-specific AU datasets and detection algorithms that can enhance the accuracy and robustness of AU recognition for this unique population.

In response to this need, we have developed an AU dataset specifically for children with autism, named the Hugging Rain Man (HRM) dataset. This dataset, manually annotated by FACS experts, covers 33 detailed action units and descriptors, aiming to capture the nuanced facial muscle movements of children with ASD across various contexts. The annotations include not only basic emotional expressions such as happiness, sadness, and anger, but also focus on unique facial features associated with ASD, such as atypical smiles \cite{Alvari2021IsST} and facial asymmetry \cite{Samad2015AnalysisOF}.

\subsection{Atypicality Rating}
To establish the relationship between AU/AD and atypical expressions, we used a five-point Likert scale \cite{likert1932technique} to annotate the atypical ratings of expression segments in the HRM dataset \cite{faso2015evaluating}. Each expression segment, from start to finish, received the same label. Due to issues like head pose, some frames were discarded during AU annotation, leading to incomplete segments. These incomplete segments were also subject to annotation. A total of five judges participated in the annotation: two FACS experts (one male and one female), one male student, and two experienced mothers. All five judges specialize in areas involving  children with ASD and have a solid understanding of their behavioral characteristics. The latter three judges performed the annotation in a double-blind manner, unaware of the group to which each participant belonged, while the FACS experts did not, as they already knew the group assignments from previous AU annotations.

The scale question was: "When you watch this facial expression segment, to what extent does your first impression suggest it is abnormal, rare, atypical, or discomforting?" The scale included five levels: 1. Natural/Normal, 2. Fairly Natural/Normal, 3. Uncertain, 4. Fairly Abnormal/Atypical, and 5. Highly Abnormal/Atypical. We calculated Fleiss' Kappa and ICC (3,1) to evaluate the consistency and reliability among judges. The results indicated moderate agreement on categorical choices (Kappa = 0.503, $P < 0.001$), whereas the consistency on numerical ratings was relatively high (ICC = 0.761, $P < 0.001$). Although differences were noted in categorical choices for some items, the judges showed good agreement on specific numerical ratings. Overall, the consistency among the five judges was at a moderate to good level. Given that "what is atypical" is highly subjective, different standards may have influenced their perceptions and annotations, leading to variability in agreement—an unavoidable factor in this study.

Finally, we averaged the atypicality ratings from the five judges to obtain a single label for each segment, thereby reducing bias due to individual subjectivity \cite{volker2009facial}.
\begin{figure*}[!htbp]
\centering
\includegraphics[width=0.95\textwidth]{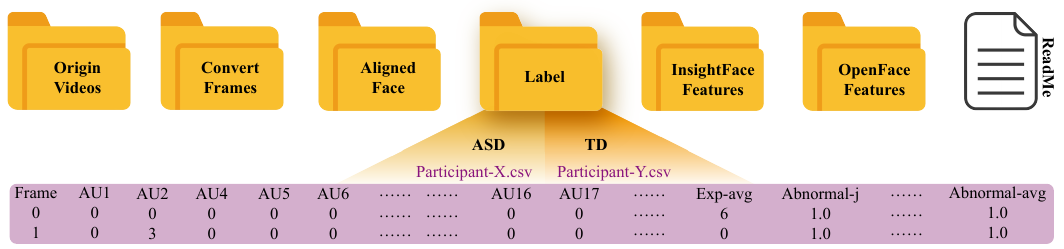}
\caption{Data organization.}
\label{fig_3_data_org}
\end{figure*}
\subsection{Data Organization}
The organization of this dataset is illustrated in Fig. \ref{fig_3_data_org}. It consists of six main folders: the "Origin Videos" folder contains the original videos for the entire dataset; the "Convert Frames" folder includes the corresponding frame images from the original videos; the "Aligned Face" folder contains face images aligned using five facial keypoints; the "Label" folder stores the annotations for all participants, including AU/AD labels, expression types, and atypicality ratings; the "InsightFace Features" folder features various automatically extracted information from the InsightFace library, such as five facial keypoints, head pose, and face bounding boxes. Similarly, OpenFace provides comparable features. Each of the six main folders contains two subfolders labeled ASD and TD, with each subject type folder housing multiple independent folders for individual participants, e.g., ASD/Participant-X. For the convenience of the research team, the ReadMe file includes additional details, such as data normalization procedures.

\section{Dataset Analysis\label{sec4}}
\subsection{AU/AD Distribution}
We begin by discussing the reasoning behind the comprehensive annotation of this dataset. Compared to traditional facial expression datasets, this study adopted a more extensive labeling approach, covering a wide range of AUs as well as several ADs related to emotional or psychological states. In addition to annotating common AUs like AU6 and AU12, we specifically focused on actions that might reflect psychological states, such as AD32 (lip biting) and AU28 (lip suck). These ADs could indicate heightened psychological tension or anxiety in  children with ASD under certain conditions \cite{PopJordanova2019DifferentCE}, subtleties that are often overlooked in conventional expression analysis. Therefore, this detailed annotation not only provides richer data for facial expression research but also offers new insights into the atypical emotional expressions of  children with ASD.

As seen in Table \ref{au_dist_reli}, we also observed a significant variation in the frequency of certain AUs/ADs. The most frequently observed AU was AU25 (lips part), which was labeled in approximately 80,000 frames, making it one of the most common actions in the dataset. Similarly, AU26/27 (jaw drop) appeared frequently, with around 50,000 frames. In contrast, other facial actions such as AU12 and AU7 also had relatively high occurrences. However, certain AUs/ADs were marked much less frequently. For instance, AU15 (lip corner depressor) appeared in only 684 frames, while AU9 (nose wrinkler) was noted in 1,924 frames. This imbalance may partly be due to the design of the experimental stimuli and ethical considerations. During task design, we avoided using stimuli that could directly evoke crying, fear, or extreme emotions in children, which resulted in a low frequency of emotionally intense AUs such as AU9 and AU15. Additionally, some AUs/ADs are less common in everyday facial expressions, such as AD35 (Suck).

This disparity in the distribution of AUs/ADs has a direct impact on training AU detection models. With actions like AU25 and AU26/27 dominating a large portion of the data, models might preferentially learn these frequent AUs during training, leading to diminished recognition performance for rarer AUs. Such class imbalance could impair the generalization of the model. Therefore, in the subsequent model training, strategies such as weighted loss functions or data augmentation techniques will be needed to address the imbalance, ensuring that the model learns all facial action units more equitably.

\subsection{Differences in Static Facial Expressions Between ASD and TD Groups}

In this section, we conducted a detailed analysis of basic facial expressions. To begin, we generated pseudo-labels for seven basic emotions using three state-of-the-art facial expression recognition algorithms: POSTER++ \cite{Mao2023POSTERVA}, EAC \cite{Zhang2022LearnFA}, and DDAMFN++ \cite{Zhang2023ADA}. By applying a soft voting method, we integrated the probability distributions from each algorithm to produce an average probability distribution for each image. A relatively lenient confidence threshold (0.6) was applied, and frames in which the highest probability exceeded this threshold were labeled as “confident” expressions.

To enhance the accuracy of emotion classification, we incorporated the established relationships between emotions and facial action units for further refinement \cite{Du2014CompoundFE,ekman1978facial}. For the "confident" surprise category, an expression is classified as surprise only when at least one of the following AUs is activated: AU1, AU2, AU5, AU25, or AU26/27. Similarly, the constraint for happiness is set to the activation of AU12. For sadness, the constraints include the activation of AU1, AU4, AU6, AU15, or AU17.

After this filtering process, we identified three frequently occurring emotions in the dataset: happiness, surprise and sadness. Based on these emotions, we conducted Mann-Whitney U tests to compare AU activation differences between the ASD and TD groups. For each AU, we reported the \textit{Z}-values and \textit{P}-values to determine whether the observed activation differences were statistically significant.

\begin{table}[!htb]
\caption{Significant Differences in AUs Between ASD and TD Groups for Three Expressions (Mann-Whitney U Test)}
\centering
\resizebox{\linewidth}{!}{
\begin{tabular}{c m{1cm}<{\centering} m{1cm}<{\centering} m{1cm}<{\centering} m{1cm}<{\centering} m{1cm}<{\centering} m{1cm}<{\centering} m{1cm}<{\centering}}
    \toprule
    \multirow{2}*{AU/AD} & \multicolumn{2}{c}{Happy} & \multicolumn{2}{c}{Surprise} & \multicolumn{2}{c}{Sad}\\ \cline{2-7}

    \rule{0pt}{9pt}&$P$ &   $Z$  &  $P$ &  $Z$ &  $P$ & $Z$ \\
    \midrule
    1 & -0.987 & 0.324 &-4.248 & \textless0.001&-4.185 &\textless0.001  \\ 
        2 & -0.966 & 0.334 & -3.789 &\textless0.001& -3.638 & \textless0.001\\ 
        4 & -5.690 &\textless0.001 &-1.597 & 0.110& -10.501 & \textless0.001\\
        5 & -1.494 & 0.135 &-2.454 & 0.014& 0& 1\\ 
        6 & -29.493 & \textless0.001 &-3.422 & 0.001& -2.335 & 0.020 \\ 
        7 & -47.485 & \textless0.001 &-1.300 & 0.194&-16.307 & \textless0.001\\ 
        43E & -8.739 & \textless0.001 &-2.536 & 0.011 & -0.849 & 0.396\\ 
        8 & -5.731 &\textless0.001 &-0.828 & 0.408& -0.457 & 0.647\\ 
        9 & -15.800 & \textless0.001 & 0& 1 & -9.289 & 0.000\\ 
        10 & -8.700 & \textless0.001 & -11.817 & \textless0.001&-4.441 &\textless0.001\\ 
        12 & -2.236 & 0.025 &-1.014 & 0.310&-4.332 & \textless0.001\\ 
        13 & -14.600 & \textless0.001 &0 & 1& 0& 1\\ 
        14 & -10.081 &\textless0.001 &-1.436 & 0.151&-3.566 & \textless0.001\\ 
        15 & -2.796 & 0.005 & 0& 1& -0.647 & 0.518\\ 
        16 & -25.609 &\textless0.001 &-4.404 &\textless0.001&-4.767 &\textless0.001\\ 
        17 & -16.982 &\textless0.001 & -2.356 & 0.018& -24.030 & \textless0.001\\ 
        18 & -0.400 & 0.689 & -1.904 & 0.057 & -2.238 & 0.025\\ 
        19 & -2.287 & 0.022 &-8.048 &\textless0.001& -1.638 & 0.101\\ 
        20 & -7.745 & \textless0.001 & -0.585 & 0.558&-5.485 & \textless0.001\\ 
        22 & -2.362 & 0.018 &-2.501 & 0.012& 0& 1\\ 
        23 & -16.362 &\textless0.001 & -2.121 & 0.034&-2.603 & 0.009\\ 
        24 & -4.732 & \textless0.001 &0 & 1&-14.603 & \textless0.001 \\ 
        25 & -6.243 & \textless0.001 &-1.858 & 0.063& -23.688 &\textless0.001 \\ 
        2X & -9.824 & \textless0.001 &-1.486 & 0.137& -13.343 & \textless0.001 \\ 
        28 & -15.550 & \textless0.001 &-0.585 & 0.558& -1.325 & 0.185\\ 
        29 & -11.677 &\textless0.001 &-0.926 & 0.355&-2.013 & 0.044\\ 
        30 & -4.611 &\textless0.001 &-3.356 & 0.001 & 0& 1\\ 
        32 & -16.789 &\textless0.001 &0 &1&-0.457 & 0.647\\ 
        33 & 0 & 1 &0 & 1& 0& 1\\ 
        34 & 0 & 1 & 0& 1& -0.264 & 0.792\\ 
        35 & 0 & 1 & 0& 1& 0& 1\\ 
        36 & 0 & 1 &-2.738 & 0.006& 0& 1\\ 
        37 & -4.992 & \textless0.001&-6.849&\textless0.001&-0.877 & 0.380\\
    \bottomrule
\end{tabular}
}
\label{u_test}
\end{table}
Across these three emotions, several AUs showed significant differences between the groups ($\textit{P} < 0.05$), as presented in Table \ref{u_test}. For the “happy” expression, AU6 ($\textit{Z} = -29.493, \textit{P} < 0.001$) and AU7 ($\textit{Z} = -47.485, \textit{P} < 0.001$) demonstrated highly significant differences, while AU12 ($\textit{Z} = -2.236, \textit{P} = 0.025$) also showed a notable difference. Significant differences were found for AU1 ($\textit{P} < 0.001$) and AU2($\textit{P} < 0.001$) in the “surprised” expression, AU1($\textit{P} < 0.001$), AU4 ($\textit{P} < 0.001$), and AU17 ($\textit{P} < 0.001$) in the “sad” expression.

\begin{figure*}[!htbp]
\centering
\includegraphics[width=0.95\textwidth]{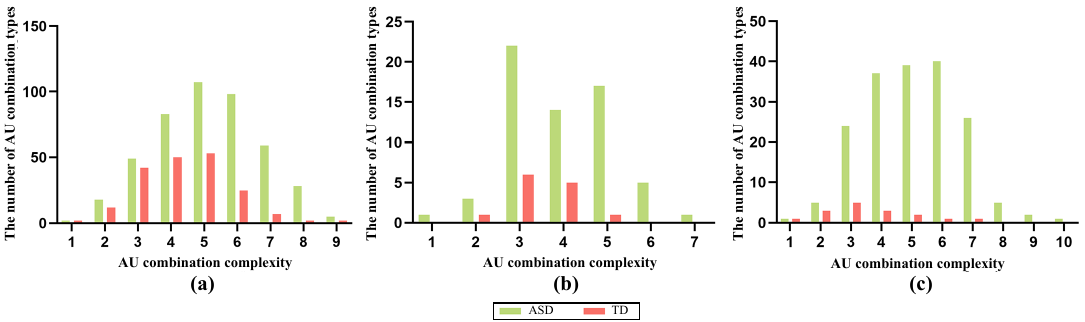}
\caption{The number of combination types at different AU combination complexity. (a) happy, (b) surprise, (c) sad.}
\label{fig_3_exp_au_combine}
\end{figure*}

To further investigate differences in AU combinations between the ASD and TD groups, we calculated the length, or AU combination complexity, of all non-repeated AU combinations for each emotion. We also counted the number of distinct AU combination types at different levels of complexity (accounting for AU direction). As illustrated in Fig.\ref{fig_3_exp_au_combine} the X-axis represents AU combination complexity, while the Y-axis reflects the number of unique AU combination types.

The data reveal that AU combination complexity in the ASD group was more diverse. For example, under the "sad" emotion, some combinations involved up to 10 distinct AUs. The ASD group frequently displayed combinations with complexities of 5, 6, and 7 AUs, suggesting more intricate facial muscle movement patterns compared to the TD group \cite{Krishnappababu2021ExploringCO}. At the same complexity level, the ASD group exhibited a greater variety of AU combinations. This higher diversity in AU combination types suggests that individuals with ASD  express emotions through a wider range of facial configurations, potentially resulting in more complex AU patterns.

Conversely, the TD group tended to rely on shorter and simpler AU combinations, particularly within the complexity range of 1 to 3 AUs. In contrast, the ASD group showed a higher number of AU combination types at greater complexity levels, typically involving 4 or more AUs. This may indicate that the facial expressions of children with TD are more standardized, with greater clarity in muscle movements and fewer complex combinations, making their expressions easier to recognize.

In summary, the ASD group exhibited not only greater AU combination complexity but also a wider variety of combinations at each complexity level. This suggests that their facial expression patterns may be more irregular and varied, possibly due to different neural mechanisms or strategies in emotional expression compared to children with TD.

\section{Benchmark Performance\label{sec5}}
Based on the statistical test results in Table \ref{u_test}, along with commonly observed AUs/ADs associated with facial expressions \cite{Du2014CompoundFE} and the distribution of each AU/AD in Table \ref{au_dist_reli} , we selected 22 AUs/ADs for both the atypicality ratings regression and the AU detection task. These include AUs/ADs: AU1, AU2, AU4, AU6, AU7, AU9, AU10, AU12, AU14, AU15, AU16, AU17, AU18, AD19, AU20, AU23, AU24, AU25, AU2X (AU26/27), AU28, AD32, and AU43E.
\subsection{Atypicality Ratings Regression}
We aim to establish the relationship between AU/AD time series and expression atypicality ratings by employing two temporal models: GRU \cite{Chung2014EmpiricalEO} and BiLSTM \cite{Hochreiter1997LongSM}, to construct baseline models. 
\subsubsection{\textbf{Cross-validation Setting}}
During model training, we implemented a subject-independent three-fold cross-validation strategy to ensure the generalization of the model. Furthermore, to validate the stability of the outcomes, we conducted 10 independent random three-fold cross-validations.
\subsubsection{\textbf{Evaluation Protocol}}
To comprehensively assess the performance of the regression task, we selected four commonly used evaluation metrics: Mean Absolute Error (MAE), Mean Squared Error (MSE), Root Mean Square Error (RMSE), and Mean Absolute Percentage Error (MAPE). These metrics will aid in gaining deeper insights into the models' predictive capabilities regarding expression abnormality ratings.
\subsubsection{\textbf{Implementation Details}}
The ratings from all five judges will be averaged to mitigate noise caused by individual biases. For the creation of AU/AD time series data, we employed a sliding window approach to construct data pairs. The time window size was empirically set to 15 frames (0.5 seconds). Using a larger time window could result in the loss of some segments with an average score greater than 2. Segments shorter than this time window were discarded to ensure data consistency and completeness. 

For the GRU and BiLSTM regression models, we added a linear layer to output predicted values and employed a dropout probability of 0.5 to enhance the models' generalization capability. The hidden layer dimension of the models is set to 128. The loss function utilized was Huber loss ($\delta=1.0$), which combines the advantages of MSE and MAE, making it more robust to outliers. We used the AdamW optimizer \cite{loshchilov2017decoupled} with an initial learning rate of $1e^{-6}$ and a scheduler that decreases the learning rate when the model fails to improve over two consecutive epochs. Model training was conducted with a batch size of 64, and the training duration was set to 15 epochs to ensure stability and optimization of model performance.

\subsubsection{\textbf{Results Analysis}}
\begin{table}[!t]
\caption{Facial expression atypicality ratings regression results}
\centering
\resizebox{\linewidth}{!}{
\begin{tabular}{ccccc}
\toprule
 & MAE & MSE & RMSE & MAPE\\
\midrule
GRU & 0.3024$\pm$2.99E-5 &0.3799$\pm$7.21E-5 & 0.3379$\pm$2.99E-5 & 0.1818$\pm$2.18E-5 \\
BiLSTM & 0.3218$\pm$4.03E-6 &0.3980$\pm$1.55E-5 & 0.3682$\pm$1.76E-6& 0.1995$\pm$2.54E-6 \\
\bottomrule
\end{tabular}
}
\label{exp_ab_regres}
\end{table}

As can be seen from Table \ref{exp_ab_regres}, GRU consistently outperforms BiLSTM, particularly in key indicators such as MAE and RMSE. This suggests that the GRU model provides more accurate predictions of atypicality ratings in this task. Given that the label range is within $[1.0, 5.0]$, the values of MAE, MSE, RMSE, and MAPE fall within a reasonable range. At this level of error, the predictive performance of both models can be considered acceptable. Thus, using temporal regression models, we have built a bridge between subjective perception and objective facial features, providing a new perspective for understanding emotional expression in individuals with ASD. This approach not only captures the dynamic changes in facial AUs but also effectively predicts the degree of atypicality in emotional expressions, offering a powerful tool for analyzing atypical expressions in ASD.

\subsection{AU Detection}
In this section, we re-implemented four open-source algorithms and validated them on the HRM dataset. These algorithms include EmoFAN \cite{Toisoul2021EstimationOC}, ME-Graph \cite{Luo2022LearningME}, MAE-Face \cite{Ma2022FacialAU}, and FMAE \cite{Ning2024RepresentationLA}. The first two (EmoFAN and ME-Graph) are supervised learning algorithms, while the latter two (MAE-Face and FMAE) are self-supervised learning algorithms. We present the results of these algorithms as benchmark performances on the HRM dataset and include ResNet-50 \cite{He2015DeepRL} as an additional baseline model. By thoroughly evaluating the performance of these models on the HRM dataset, we aim to provide reliable benchmarks for future research in this area.


EmoFAN utilizes facial landmarks to extract emotion-related features, assisting in both facial expression recognition and arousal-valence evaluation. Low-level and high-level features extracted by the Facial Alignment Network (FAN) \cite{Bulat2017HowFA} are used to compute attention maps, guiding the network to focus on emotion-relevant facial regions. For AU detection, we adapted EmoFAN by replacing the emotion classification/regression task with an AU detection task.

ME-Graph models the relationships between multiple AU vertex features. These vertices represent AU activation states and their associations with other AUs. The feature maps extracted from the backbone, along with these vertex features, are used to construct an AU relationship graph with multi-dimensional edge features. This graph is processed by a Graph Convolutional Network (GCN) \cite{Bresson2017ResidualGG} to detect AUs.

MAE-Face performs a masking-then-reconstruction task \cite{He2021MaskedAA} using approximately 200 million images sourced from AffectNet \cite{Mollahosseini2017AffectNetAD}, CASIA-WebFace \cite{Yi2014LearningFR}, IMDB-WIKI \cite{Rothe2015DEXDE}, and CelebA \cite{Liu2014DeepLF} datasets. The pre-trained Vision Transformer (ViT) \cite{Dosovitskiy2020AnII} encoder is fine-tuned in an end-to-end manner specifically for AU detection, allowing the model to leverage its learned representations for more accurate action unit recognition.

FMAE is trained on an even larger corpus of facial images, roughly 900 million, to build a more robust MAE model \cite{He2021MaskedAA}. To improve generalization in AU detection and remove identity-related biases, FMAE incorporates Identity Adversarial Training (IAT) to learn identity-invariant features, referred to as FMAE-IAT.

\subsubsection{\textbf{Cross-validation Setting}}
We employed a 3-fold subject-independent cross-validation approach to split the training and test sets \cite{Shao2020JANetJF,Luo2022LearningME,Ma2022FacialAU,Yang2023TowardRF}. The AU distribution was maintained similarly across all folds. For each fold, two-thirds of the data were used for training, while the remaining one-third served as the test set. Detailed information on the AU distribution for each fold can be found in Table \ref{AU_distribution}.

\begin{table*}[!htbp]
\caption{AU distribution in three-fold cross-validation}
\centering
\resizebox{\linewidth}{!}{
\begin{tabular}{cccccccccccccccccccccccc}
\toprule
    &1 & 2 & 4 & 6 & 7 & 9 & 10 & 12 & 14 & 15 & 16 & 17 &18 &19& 20 & 23 & 24 & 25 & 26/27 &28 &32 &43E & Frame \\ \midrule
    Fold-1&5,565 & 4,499 & 5,570 & 3,179 & 6,027 & 459 & 3,146 & 9,429 & 7,260 & 272 & 1,906 & 2,289&572&387 & 865 & 2,132 & 2,642 & 24,753 & 15,030&637&375&1,569 & 44,849 \\ 
    Fold-2&6,262 & 4,624 & 4,133 & 5,029 & 6,840 & 755 & 1,590 & 11,734 & 5,227 & 254 & 2,866 & 2,500&2,049&1,864 & 1,418 & 1,975 & 1,557 & 31,430 & 19,989&1,587&1,215&918 & 45,487 \\ 
    Fold-3&6,012 & 4,995 & 4,285 & 4,337 & 6,866 & 711 & 2,188 & 10,460 & 6,450 & 158 & 2,814 & 1,597&819&521 & 1,587 & 1,983 & 1,769 & 24,500 & 16,163&1,616&448&816 & 41,422 \\ 
    \bottomrule
\end{tabular}
}
\label{AU_distribution}
\end{table*}

\subsubsection{\textbf{Evaluation Protocol}}
To evaluate the performance of the aforementioned algorithms, we used two commonly employed metrics in AU detection: Accuracy and F1-Score. Accuracy represents the proportion of correctly identified samples out of the total samples. While it is a straightforward measure, it may not fully reflect the classifier’s performance, especially in scenarios with imbalanced classes. The formula for accuracy is as follows: 
\begin{equation}
\label{acc}
Accuracy (ACC)=\frac{TP+TN}{TP+TN+FP+FN}
\end{equation}
where $TP$ is true positive, $TN$ is true negative, $FP$ is false positive, $FN$ is false negetive.

F1-Score is the harmonic mean of Precision and Recall. It is particularly useful when dealing with imbalanced classes, as it balances both precision and recall. The F1-Score ranges from 0 to 1, with higher values indicating better model performance. The formula is as follows:
\begin{equation}
\label{f1}
F_1 =2 \times \frac{Precision \times Recall}{ Precision+  Recall} 
\end{equation}

\subsubsection{\textbf{Implementation Details}}
In the AU detection experiments, to ensure consistent model performance across different samples and improve training efficiency, we first normalized the entire dataset. Normalization reduces feature discrepancies caused by external factors such as lighting and shooting distance, ensuring a more stable data distribution. This is crucial for the convergence and robustness of deep learning models. To achieve this, we calculated the mean and variance of the pixel values across all images in the dataset, and each image was standardized accordingly to make the features more suitable for optimization in neural networks.

The first step of data preprocessing involved face detection using the InsighFace library \cite{Deng2018ArcFaceAA}. This tool not only accurately detects faces but also provides important facial attributes such as bounding boxes, 106 2D landmarks, 68 3D landmarks, head pose estimations (pitch, yaw, etc.), and 5 key points. These facial features may be used in subsequent processing and AU detection. By saving this data, we provided crucial prior information for other algorithms and analyses that may need it.

Using the 5 key points returned by RetinaFace \cite{deng2020retinaface} (centers of both pupils, nose tip, and mouth corners), we performed face alignment. This step adjusted faces into a unified reference coordinate system, eliminating deviations caused by different camera angles. Ensuring facial features from different angles and poses are analyzed within the same spatial context improved the accuracy and robustness of the subsequent AU detection models. 

For ResNet-50, we initialized the model with pre-trained weights from ImageNet \cite{Russakovsky2014ImageNetLS}. Data augmentation was performed using random cropping, random flipping, and color jitter. Additionally, we employed the binary cross entropy as the loss function. The final fully connected layer output was set to 22 dimensions, corresponding to the 22 AUs/ADs to be detected. The initial learning rate was set to $1e^{-4}$, and we trained the model for 20 epochs with a batch size of 64, using the AdamW optimizer and a cosine annealing schedule.

\begin{table*}[htbp]
\caption{Comparison of Supervised and Self-Supervised Methods for AU Detection}
\centering
\resizebox{\linewidth}{!}{
\begin{threeparttable}
\begin{tabular}{ccccccccccccc}
\toprule
\multirow{3}{*}[-0.2cm]{AU} & \multicolumn{6}{c}{Supervised} & \multicolumn{6}{c}{Self-Supervised} \\ \cline{2-13}
 & \multicolumn{2}{c}{ResNet-50 \cite{He2015DeepRL}}  & \multicolumn{2}{c}{EmoFAN \cite{Toisoul2021EstimationOC}} & \multicolumn{2}{c}{ \makecell{ME-GraphAU \cite{Luo2022LearningME}\\(ResNet-50)}} & \multicolumn{2}{c}{FMAE \cite{Ning2024RepresentationLA}} & \multicolumn{2}{c}{FMAE-IAT \cite{Ning2024RepresentationLA}}& \multicolumn{2}{c}{MAE-Face \cite{Ma2022FacialAU}}\\ \cline{2-13}
 \rule{0pt}{9pt}& ACC & F1 & ACC & F1 & ACC & F1 & ACC & F1 & ACC & F1 & ACC &F1 \\ 
 \midrule
1 & 87.52  & 48.64  & 84.40  & 50.60  & 86.30  & 53.15  & \underline{89.01}  & 53.26  & 88.34  & 54.10  & 88.99  & \textbf{55.00} \\ 
        2 & 89.29  & 48.49  & 85.40  & 44.83  & 88.42  & 49.97  & 91.10  & 52.46  & 90.99  & \textbf{55.79} & \underline{91.34}  & 53.74  \\ 
        4 & 93.24  & 60.30  & 91.53  & 64.67  & 92.09  & 62.16  & \underline{94.38}  & 70.06  & 93.35  & 63.27  & 93.92  & \textbf{70.08}  \\ 
        6 & 92.49  & 58.82  & 90.13  & 57.47  & 91.75  & 57.49  & \underline{93.34}  & \textbf{62.60}  & 93.17  & 61.49  & 93.30  & 60.51  \\ 
        7 & 90.76  & 66.68  & 87.10  & 64.77  & 89.29  & 65.51  & 90.85  & \textbf{68.03} & \underline{91.11}  & 66.82  & 90.88  & 67.15  \\ 
        9 & 98.53  & 35.92  & 97.27  & 34.17  & 98.74  & 51.73  & 98.76  & 50.60  & 98.93  & \textbf{56.10}  & \underline{98.95}  & 53.80  \\ 
        10 & \underline{92.92}  & 20.96  & 87.37  & 29.53  & 92.26  & \textbf{30.75}  & 92.53  & 27.08  & 92.20  & 28.32  & 92.35  & 25.36  \\ 
        12 & 90.81  & 79.74  & 89.40  & 79.63  & 90.06  & 80.11  & 64.01  & 83.02  & 91.99  & 83.24  & \underline{92.41}  & \textbf{84.19}  \\ 
        14 & 89.28  & 55.41  & 85.17  & 51.90  & 87.98  & 55.66  & 90.13  & 61.06  & 89.80  & 62.16  & \underline{90.18}  & \textbf{62.68}  \\ 
        15 & 99.39  & 10.58  & 99.13  & 15.07  & \underline{99.45}  & \textbf{27.93}  & 99.43  & 9.91  & \underline{99.45}  & 0.00  & 99.43  & 3.29  \\ 
        16 & 93.02  & 23.15  & 88.93  & \textbf{28.53}  & 93.04  & 28.15  & 93.10  & 17.17  & 92.55  & 16.53  & \underline{93.61}  & 23.80  \\ 
        17 & 95.46  & 40.10  & 91.47  & 38.40  & 95.04  & 45.47  & 96.10  & 50.10  & 95.99  & 47.99  & \underline{96.26}  & \textbf{50.14}  \\ 
        18 & 97.67  & 35.89  & 95.57  & 37.63  & 97.39  & 39.53  & 97.25  & 42.59  & \underline{97.89}  & \textbf{47.92}  & 97.67  & 43.27  \\ 
        19 & 98.14  & 30.56  & 97.20  & 36.27  & 98.46  & 49.62  & 98.53  & 47.30  & 98.71  & 57.53  & \underline{98.96}  & \textbf{65.23}  \\ 
        20 & 97.12  & 26.31  & 95.73  & 33.17  & 97.21  & 37.85  & \underline{97.67}  & \textbf{50.90}  & 97.65  & 49.42  & 97.54  & 46.32  \\ 
        23 & 95.03  & 22.43  & 90.20  & 27.27  & 94.24  & \textbf{29.23}  & \underline{95.14}  & 20.80  & 94.70  & 24.01  & 94.71  & 24.19  \\ 
        24 & 95.29  & 33.34  & 92.80  & \textbf{39.63}  & 94.89  & 39.36  & 95.41  & 38.93  & \underline{95.58}  & 37.06  & 95.57  & 33.69  \\ 
        25 & 90.69  & 92.30  & 91.00  & 92.50  & 89.58  & 91.64  & 91.81  & 93.24  & 92.22  & 93.51  & \underline{92.87}  & \textbf{94.13}  \\ 
        2X & 76.74  & 68.36  & 77.50  & 71.63  & 75.57  & 69.21  & 78.67  & \textbf{71.68}  & 78.06  & 70.06  & \underline{78.76}  & 71.51  \\ 
        28 & 97.46  & 43.44  & 96.03  & 47.07  & 97.52  & 50.71  & 97.35  & 44.63  & 97.23  & 41.32  & \underline{97.58}  & \textbf{49.41}  \\ 
        32 & 98.79  & 29.78  & 98.10  & 39.97  & \underline{98.82}  & 39.32  & 98.64  & 31.14  & 98.81  & \textbf{42.04}  & 98.61  & 40.54  \\ 
        43E & 97.95  & 65.43  & 96.67  & 57.43  & 97.84  & 67.09  & 98.57  & 71.24  & 98.54  & 69.56  & \underline{98.68}  & \textbf{72.08}  \\ 
        Avg. & 93.53  & 45.30  & 91.28  & 47.37  & 93.00  & 50.98  & 92.81  & 50.81  & 93.97  & 51.28  &\underline{94.21}  & \textbf{52.28} \\ 
\bottomrule
\end{tabular}
\begin{tablenotes}
        \footnotesize
        \item[1] The optimal ACC and F1 scores are indicated by underlining and bolding, respectively.
      \end{tablenotes}
\end{threeparttable}
}
\label{AU_detection_resu}
\end{table*}
For EmoFAN, data augmentation involved random horizontal flipping, and the loss function was the same as JÂA-Net's weighted multi-label cross-entropy loss \cite{Shao2020JANetJF}. Additionally, the samples were weighted based on the ratio of positive to negative instances for each AU\footnote{\url{https://github.com/jingyang2017/aunet_train}}. The optimizer used was AdamW with an initial learning rate of $1e^{-4}$, and the model was trained for 12 epochs with a batch size of 64.

For ME-Graph, we report the results using ResNet-50 as the backbone, allowing for direct comparison with the basic ResNet-50 model. Data augmentation was inspired by JÂA-Net, with additional color jittering, and images were random cropped to 224$\times$224. The initial learning rate was set to $1e^{-4}$, using a weighted asymmetric loss function, and the optimizer was AdamW with $\beta_1 = 0.9$, $\beta_2 = 0.999$, and a weight decay of $5e^{-4}$. The training batch size was 64, nearest neighbor setting for the Facial Graph Generator (FGG) module is set to 4.

For MAE-Face and FMAE, we directly employed the ViT-Large-based pre-trained models provided by the author\footnote{\url{https://github.com/forever208/FMAE-IAT}} for end-to-end fine-tuning on the HRM dataset, maintaining all parameters at their default settings, except for the learning rate, batch size and epochs. The base learning rate for both MAE-Face and FMAE was set to $2e^{-4}$. For FMAE-IAT, gradient reversal was used for identity adversarial training, with an initial learning rate set to $1e^{-3}$. All three self-supervised algorithms were trained for 30 epochs with a batch size of 32, employing AutoAugmentation \cite{cubuk2019autoaugment} and random erasing while disabling mixup.

The supervised algorithms were trained from scratch, while the self-supervised algorithms employed an end-to-end fine-tuning approach \cite{Ning2024RepresentationLA}.

\subsubsection{\textbf{Results Analysis}}
The comparison results of all six algorithms are listed in Table \ref{AU_detection_resu}. In the self-supervised, end-to-end fine-tuning experiments for AU detection, MAE-Face achieved the highest F1-score and accuracy. Compared to FMAE and FMAE-IAT, which also utilized masked autoencoding and were trained on approximately 900 million facial images with a ViT-Large architecture, MAE-Face outperformed them by 1\% to 2\%, despite being trained on a smaller dataset of about 200 million facial images and using a ViT-Base architecture. This performance advantage may be attributed to MAE-Face's use of a ViT encoder pretrained on ImageNet-1k, providing a strong initialization for subsequent tasks, whereas the FMAE series models were trained from scratch. Furthermore, FMAE-IAT achieved the second-best results on both metrics. It employed Identity Adversarial Training (IAT), which uses a gradient reversal approach to learn identity-independent features, effectively preventing the model from capturing identity-related information. This strategy was a key reason for its superior performance compared to FMAE.

Among the supervised algorithms, ME-GraphAU achieved the highest F1-score and accuracy, owing to its innovative graph structure learning approach. This method explicitly models the relational cues between pairs of AUs in facial expressions, encoding both node features and multidimensional edge features within a graph structure. This not only enhances the representation of AU relations but also enables the model to capture complex AU interactions, thereby improving the accuracy of AU recognition. The multi-level features and landmark-based heatmap attention mechanism allow EmoFAN to slightly outperform the baseline (ResNet-50) in terms of F1-score. EmoFAN leverages a pretrained FAN to directly predict facial landmarks and generate heatmaps as attention maps, which are then applied to multi-level (low-level and high-level) facial features. This approach provides enhanced spatial focus and contextual awareness, thereby improving the accuracy of AU prediction.

Overall, the performance of self-supervised algorithms reached results comparable to those of supervised algorithms, with some even surpassing their supervised counterparts. Self-supervised algorithms leverage vast amounts of data to help models learn richer and more general facial feature representations, whereas supervised methods often rely heavily on labeled data, which is scarce and costly for AU detection tasks \cite{Ning2024RepresentationLA}. The limited availability of annotated data may hinder the model's ability to learn comprehensive features and could even lead to overfitting on identity. Therefore, large-scale self-supervised pretraining can compensate for the lack of labeled data. Specifically, self-supervised learning, particularly MAE, learns to focus on facial details and texture information during the reconstruction process, which is crucial for AU detection that involves fine-grained facial muscle movements. This ability to learn fine details is one of the key advantages.

\section{Discussion\label{sec_dicus}}
Open-source AU detection tools have demonstrated excellent performance in adult facial expression analysis \cite{namba2021assessing}. However, their generalization capability is notably limited when applied to  children with ASD. As observed from the F1-scores in Tables \ref{au_dist_reli} and \ref{AU_detection_resu}, the performance of the pretrained model based on an adult facial dataset without fine-tuning is significantly lower than that of the ResNet-50 baseline. This suggests that it is not always appropriate to directly apply pre-trained models based on adult expressions to  children with ASD, whether implicitly or explicitly assuming that the facial expression features of adults and  children with ASD are similar \cite{Li2021ATM}. The study by Witherow et al. also had similar findings \cite{witherow2023deep}. Such limitations may stem from differences in facial structure, expression performance, and muscle movement patterns between children and adults \cite{Grossard2020ChildrenWA}, making it difficult for models pre-trained on adult data to accurately capture the unique AU activation patterns of children. The achievement of more accurate AU detection is attributed to our newly constructed AU dataset, which encompasses a diverse range of facial expressions from both  children with ASD and children with TD, as well as the excellent AU detection algorithms proposed in prior studies \cite{Shao2020JANetJF,Toisoul2021EstimationOC,Luo2022LearningME,Ma2022FacialAU,Ning2024RepresentationLA}. This ultimately prompted us to share this advanced model with the research community. Future research could focus on refining the annotation of AU intensity, as understanding the varying degrees of expression can provide deeper insights into emotional states and behavioral responses.


In the statistical analysis of static images, our research identified significant differences in AU activation patterns between  children with ASD and children with TD across three basic emotions (happiness, surprise, and sadness). Notably, in the happiness expression, several AUs (e.g., AU6, AU7, AU12) exhibited marked intergroup differences. This suggests that the actions of facial muscles in expressing emotions among  children with ASD may significantly differ from children with TD, supporting prior research conclusions regarding the atypical nature of facial expressions in children with ASD \cite{Webster2021ReviewPV,Grossard2020ChildrenWA}. Furthermore, in the analysis of AU combination complexity across the three emotions,  children with ASD exhibited higher AU combination complexity compared to their TD peers, and at the same level of AU complexity, they displayed a more diverse set of AU combination patterns. This suggests that their facial expression patterns may be more irregular and diverse, which could be one of the reasons why their expressions are perceived as more ambiguous \cite{Grossard2020ChildrenWA}, harder to interpret \cite{Brewer2015CanNI}, and atypical \cite{Guha2015OnQF}.

Studies evaluating the naturalness/atypicality of facial expressions in individuals with ASD have investigated how observers perceive these expressions, whether posed or spontaneous \cite{Macdonald1989RecognitionAE, volker2009facial ,faso2015evaluating}. The naturalness/atypicality of expressions was typically assessed using rating scales, which is also the approach we adopted. These studies revealed that, compared to TD individuals, expressions from ASD individuals are perceived as less natural, but the specific objective facial features that contribute to this perception have not been thoroughly explored \cite{Briot2021NewTA}. Our work extends these findings by incorporating both subjective ratings and objective FACS features to evaluate the atypicality of facial expressions. We use regression algorithms based on dynamic AU sequences to bridge the gap between subjective perception and objective features, providing a more comprehensive understanding of the atypical facial expressions in individuals with ASD.

In the dynamic sequences regression of atypicality, we achieved only a moderate to good level of consistency for atypical expression annotations (Kappa = 0.565, ICC = 0.761), this may be due to the inherently subjective nature of the 'atypical' concept and the varying interpretations of atypicality among judges. Additionally, judges may have also been influenced by the overall appearance of participants with ASD rather than the strangeness of the expressions itself \cite{volker2009facial}. The use of untrained judges may be a limitation; however, their evaluations might more closely resemble the assessments people make during everyday social interactions, which could be more socially relevant \cite{juslin2005new}. To mitigate individual bias and noise, we averaged the ratings from multiple judges, which allowed our temporal model to more effectively extract correlations with atypicality ratings from AU sequences (MAE = 0.3024, MSE = 0.3799). This approach indicates that the temporal features contained in AU sequences hold significant potential for the automatic detection of atypical expressions. Even in contexts with high subjective annotation, the temporal model can still identify patterns. We hypothesize that this is because AU sequences reflect subtle facial muscle changes, which can correlate with the judges' perceptions of atypical expressions. Thus, even with imperfect annotation consistency, the model can leverage these dynamic features for reasonably accurate predictions of atypicality ratings. This finding lays the groundwork for automated atypical expression analysis, particularly in research on the facial expressions of children with ASD. It demonstrates that FACS annotations provide an objective basis for predicting atypicality, supporting future efforts in ASD early screening.

Since our data originate from multiple tasks rather than a single timed task, we cannot definitively conclude similar findings, such as that  children with ASD have greater facial expression asymmetry \cite{Witherow2024PilotST} or fewer happy expressions compared to  children with TD \cite{Alvari2021IsST, Samad2019APS}. This aspect could be further explored in future research, particularly by setting tasks under identical conditions to validate this hypothesis. Nonetheless, our data and analytical methods offer advantages. By integrating expression data from multiple tasks, we can capture the diversity of expression patterns in children with ASD  (both posed and spontaneous), mitigating potential biases from a singular context. This multi-task data design enhances the generalizability of our findings, providing valuable support for more nuanced expression research in the future.

To foster the advancement of the research community, we have made available the labels, some machine-extracted features, and the pre-trained model weights for AU detection. This open sharing aims to enable other researchers to build upon our work, thereby facilitating progress in understanding and analyzing facial expressions in  children with ASD. Despite the temporary unavailability of the facial image data due to privacy concerns, future research could explore approaches to anonymize children's facial images while preserving critical emotional attributes \cite{gong2020disentangled,barattin2023attribute}, such as expressions or AU information. This could involve advanced techniques in facial obfuscation that maintain key features relevant to emotional analysis, ensuring both privacy and utility. By developing robust anonymization methods, future datasets could provide a balance between data accessibility and ethical standards, ultimately enabling more comprehensive collaboration and innovation in the field.

\section{Conclusion\label{sec_conclu}}
In this paper, we present a novel dataset, Hugging Rain Man (HRM), specifically designed for the analysis of atypical facial expressions in children with ASD. Unlike previous datasets, HRM encompasses both ASD and TD control groups, featuring facial images collected from a variety of experimental tasks that span a rich spectrum of both posed and spontaneous facial expressions. Our dataset also includes extensive AU/AD labels meticulously annotated by FACS experts, along with ratings of facial expression atypicality.

Based on these AU/AD annotations, we conducted analyses on both static images and dynamic sequences, gaining insights into the characteristics of atypical facial expressions in autism. Specifically, statistical analysis of static images revealed significant differences in multiple AUs/ADs between children with ASD and their TD peers when displaying the same emotional expressions. Furthermore, children with ASD exhibited more diverse and complex combinations of AUs/ADs. By leveraging a temporal regression model, we successfully established a connection between objective dynamic AU sequences and perceived expression atypicality, suggesting promising possibilities for rapid screening of  children with ASD based on facial expression sequences.

Although the facial images are not publicly available temporarily, we have made the accompanying annotations and pretrained models accessible to the research community. We hope that our dataset and resources will contribute to advancements in the fields of emotion analysis and ASD research. Future work will focus on enhancing our understanding of dynamic changes in facial AU intensity in relation to atypical expressions in ASD. Additionally, we will explore methods for anonymizing children's faces while preserving emotional attributes (including expressions and AUs), thereby facilitating the eventual public release of the dataset. This will further promote the use and impact of our dataset within the scientific community.

\section*{Acknowledgments}
The success of this study owes much to the dedicated efforts of those involved in the dataset creation process. We sincerely thank all the staff members who participated in data collection (in no particular order), including Wei Shen, Ying Zhang, Junlin Hu, Yating Dai, Yuanxu Jin, Chenglin Xie, Feixiang Li, Meijuan Luo, Xiaohui Wu and Mengyi Liao. Their hard work ensured the quality and integrity of the data. We would also like to extend our special thanks to the data preprocessing team, including Shujuan Zhou, Lingjian Ye, Meixing Lu, Siyan Zhang, and Suyun Tang. Their meticulous and rigorous work provided a solid data structure for this study, laying a strong foundation for subsequent experiments.

This work was supported by the National Natural Science Foundation of China under Grant 62377018, the self-determined Research Funds of CCNU from the Colleges’ Basic Research and Operation of MOE under Grants CCNU24ZZ154, and the China Postdoctoral Science Foundation under Grant No. 2023M742718.


%
\bibliographystyle{IEEEtran}
\bibliography{IEEEabrv,ref}

\vfill

\end{document}